\documentclass[final]{cvpr}

\usepackage{times}
\usepackage{epsfig}
\usepackage{graphicx}
\usepackage{amsmath}
\usepackage{amssymb}
\usepackage{color}
\usepackage{subfigure}
\usepackage{float}
\usepackage{comment}
\usepackage{booktabs}
\usepackage{multirow}
\usepackage{arydshln}
\usepackage{algorithm}  
\usepackage{algpseudocode}  
\usepackage{amsmath}

\usepackage[pagebackref=true,breaklinks=true,colorlinks,bookmarks=false]{hyperref}

\def\cvprPaperID{1395} 
\def\confYear{CVPR 2021}

\begin{document}
\title{Zero-Shot Instance Segmentation}

\author{Ye Zheng\textsuperscript{1,2} \qquad Jiahong Wu\textsuperscript{3}  \qquad Yongqiang Qin\textsuperscript{3}  \qquad Faen Zhang\textsuperscript{3} \qquad Li Cui\textsuperscript{1}\\
Institute of Computing Technology, Chinese Academy of Sciences\textsuperscript{1}\\
University of Chinese Academy of Sciences\textsuperscript{2}\qquad
AInnovation Technology Co., Ltd\textsuperscript{3}\\
{\tt\small \{zhengye@ict.ac.cn,wujiahong@ainnovation.com,qinyongqiang@ainnovation.com}
\\{\tt\small zhangfaen@ainnovation.com,lcui@ict.ac.cn\}}
}

\maketitle
\pagestyle{empty}
\thispagestyle{empty}

\begin{abstract}
Deep learning has significantly improved the precision of instance segmentation with abundant labeled data. However, in many areas like medical and manufacturing, collecting sufficient data is extremely hard and labeling this data requires high professional skills. We follow this motivation and propose a new task set named zero-shot instance segmentation (ZSI). In the training phase of ZSI, the model is trained with seen data, while in the testing phase, it is used to segment all seen and unseen instances. We first formulate the ZSI task and propose a method to tackle the challenge, which consists of Zero-shot Detector, Semantic Mask Head, Background Aware RPN and Synchronized Background Strategy. We present a new benchmark for zero-shot instance segmentation based on the MS-COCO dataset. The extensive empirical results in this benchmark show that our method not only surpasses the state-of-the-art results in zero-shot object detection task but also achieves promising performance on ZSI. Our approach will serve as a solid baseline and facilitate future research in zero-shot instance segmentation. 
Code available at \href{https://github.com/zhengye1995/Zero-shot-Instance-Segmentation}{ZSI}.
\end{abstract}
\begin{figure}[htbp]
\begin{center}
\includegraphics[width=\linewidth]{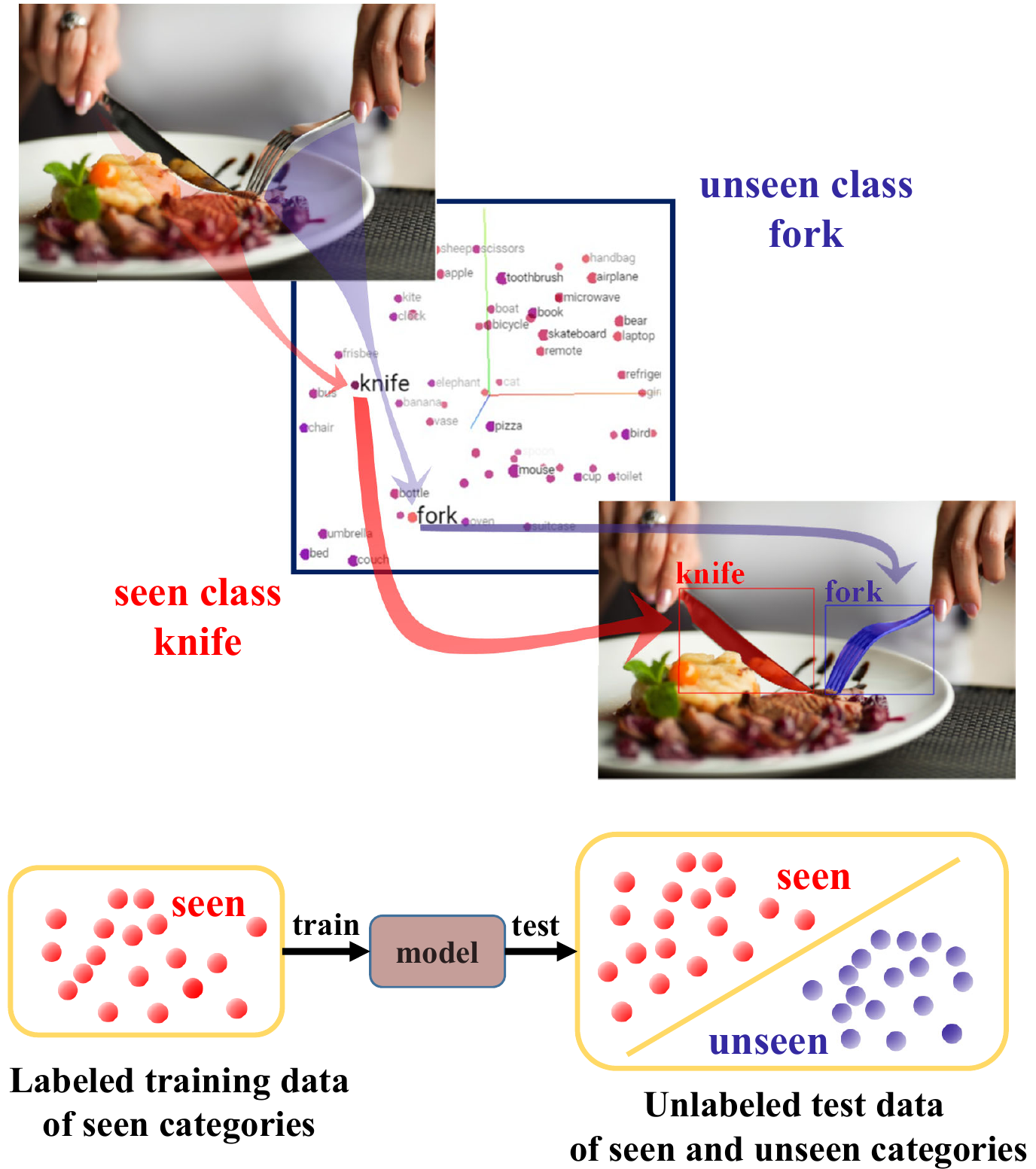}
\caption{In zero-shot instance segmentation, we can only use the labeled data of seen categories for training but predict the instance segmentation results for both seen and unseen categories. In our method, we use the seen classes data, \eg, ``knife" to establish the mapping relationship between visual and semantic concepts during training and then transfer it to segment unseen instances, \eg,``fork" in inference.}
\label{fig:zsi_example}
\end{center}
\vspace{-1cm}
\end{figure}

\section{Introduction}
\label{sec:intro}
In recent years, deep learning based instance segmentation methods~\cite{he2017mask,li2017fully,bolya2019yolact,chen2019hybrid,xie2020polarmask,wang2019solo} have made great progress. These supervised learning paradigm methods strongly rely on large-scale labeled data. However, for many real-world applications, \eg, the medical and manufacturing, collecting and labeling data are very time consuming and need professional annotators, which results in that we always do not have labeled data for unseen classes in these tasks. Besides, for open-set~\cite{bendale2016towards,yoshihashi2019classification} instance segmentation tasks, we can not label all categories, so there are many unlabeled unseen classes that need to be segmented. Without abundant labeled data, modern instance segmentation approaches are not competent to train a deep neural network to segment unseen instances. In recent years, many zero-shot learning methods have been proposed. The latest achievements of zero-shot learning focus on the problem of zero-shot classification~\cite{bendale2016towards,changpinyo2016synthesized,kodirov2017semantic,rahman2018unified,xian2017zero,zhang2015zero,zhang2016zero1,zablocki2019context,krishna2017visual,mishra2018generative,kumar2018generalized,verma2020meta}. Limited by the existing benchmarks~\cite{nilsback2008automated,russakovsky2015imagenet,welinder2010caltech}, the zero-shot classification approaches focus on reasoning a single dominant unseen object in input image, which makes it unsuitable for real scenes. In the real world, several unseen objects belonging to different classes may appear at the same time. Therefore, zero-shot object detection (ZSD)~\cite{bansal2018zero,rahman2018zero} and zero-shot semantic segmentation (ZSS)~\cite{bucher2019zero} have been proposed. ZSD is aiming to simultaneously localize and recognize unseen objects and ZSS is used to segment unseen classes, these tasks are more practical for the real-world scenarios. However, the bounding box results obtained from ZSD and the segmentation results of the entire image from ZSS are still not fine enough when we need pixel-level segmentation results of each instance. In order to meet this demand for finer results, we introduce a new problem setting, called zero-shot instance segmentation (ZSI). As illustrated in Fig~\ref{fig:zsi_example}, the goal for ZSI is not only to detect all unseen objects, but also to further precisely segment each unseen instance.

There are two main challenges for ZSI task. (i) How to do instance segmentation for unseen classes. Without unseen classes data, we can not train a deep neural network to segment unseen instances. We introduce extra semantic knowledge contained in pre-trained word-vectors to correlate the seen and unseen classes. We use the semantic word-vector and image data of seen classes to establish the visual-semantic mapping relationship in a detection-segmentation manner and transfer it from seen to unseen classes. We propose the zero-shot detector and Semantic Mask Head (SMH) to detect and segment each unseen instance. We discuss the details in Section~\ref{sec:zero-shot detector} and Section~\ref{sec:smh}. (ii) How to reduce the confusion between background and unseen classes. Unlike the zero-shot classification task that only has one domain object in each image and does not need to consider the background class, ZSI needs to distinguish foreground and background. Since the unseen data are not observed during training, the model is likely to identify the unseen objects as the background, which has a great impact on the performance. We believe that the representation of background class is the key to solve this problem. We find that the current representation of the background class has two major drawbacks: (i) The existing semantic representation of background is unreasonable. Previous works~\cite{bansal2018zero,rahman2018zero} used the word-vector for the ``background" word to represent background class. However, in computer vision task, this simple word-vector learned from unlabeled text data is not enough to represent the complex background; (ii) The existing semantic representation of the background class is fixed, which makes it difficult to adapt to the changing background in different images. To obtain a reasonable and dynamic adaptive word-vector for background class, We propose Background Aware RPN (BA-RPN) and Synchronized Background Strategy (Sync-bg). We introduce the details in Section~\ref{sec:barpn and sync-bg}.

To facilitate the research of zero-shot instance segmentation, we propose the experimental protocol and benchmark for ZSI based on the challenging MS-COCO dataset. Because instance segmentation has always been regarded as a downstream task of object detection, we set up the dataset for ZSI based on the existing ZSD benchmark in \cite{bansal2018zero,rahman2020improved}.

To summarize, the main contributions of this paper are as follows: (i) we introduce the problem of zero-shot instance segmentation in real-world settings; (ii) we propose a novel method which tackles ZSI problem in a background aware detection-segmentation manner, including Zero-shot Detector, Semantic Mask Network, Background Aware RPN and Synchronized Background Strategy. (iii) we put forward a new experimental benchmark for ZSI to evaluate the performance of the model; (iv) we provide extensive experimental and ablation studies to highlight the benefits of the proposed methods, and the results show that our method surpasses the state-of-the-art ZSD works and achieve promising performance on ZSI.
\begin{figure}[tbp]
\begin{center}
\includegraphics[width=1.0\linewidth]{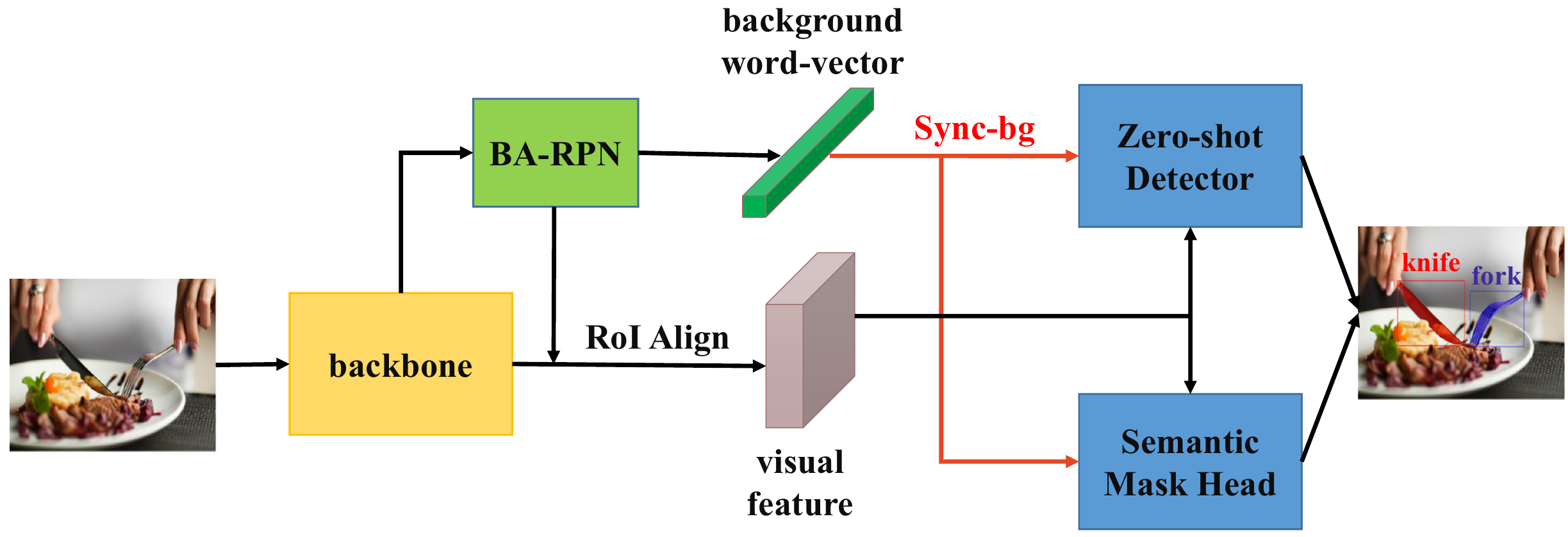}
\caption{The whole architecture for our zero-shot instance segmentation framework. For an input image, we obtain the visual feature and background word-vector for each proposal from backbone and BA-RPN through RoI Align. Then we use Sync-bg to synchronize the word-vector for background class in Zero-shot Detector and Semantic Mask Head. We can get the instance segmentation results from these structures.}
\label{fig:whole-framework}
\end{center}
\vspace{-0.5cm}
\end{figure}
\section{Related Works}
\label{sec:relatedworks}
\subsection{General Instance Segmentation}
Instance segmentation is a classical problem in the field of computer vision. Deep learning based instance segmentation methods have made great progress in the past several years, \eg, Mask R-CNN~\cite{he2017mask}, FCIS~\cite{li2017fully}, YOLCAT~\cite{bolya2019yolact}, HTC~\cite{chen2019hybrid}, PolarMask~\cite{xie2020polarmask} and SOLO~\cite{wang2019solo}. Some of them solve this problem in a detection-segmentation manner, \eg, as a typical work, Mask R-CNN adds a mask branch to Faster R-CNN which detects all objects first and then segments all pixels in each object. The above methods work in an intensive supervision manner and they are difficult to extend to novel categories without training samples.

\subsection{Zero-shot Learning}
Zero-shot learning is a classical problem in the research community of machine learning, which aims at using seen classes data to train the network and reason about unseen classes. For the past few years, several achievements have been proposed for zero-shot learning~\cite{bendale2016towards,changpinyo2016synthesized,kodirov2017semantic,rahman2018unified,xian2017zero,zhang2015zero,zhang2016zero1,zablocki2019context,krishna2017visual,mishra2018generative, kumar2018generalized,verma2020meta}. Most of them have employed the transfer learning method that transfers the knowledge learned from the seen classes to the unseen classes through various intermediate representations, \eg, the semantic word-vectors learned with unsupervised fashion from unannotated text data and the attributes of manual design. However, these methods focus on zero-shot classification task which only recognizes one domain object of the input image, and it is far from the real world.

In recent years, some work of ZSD have been reported. Rahman \etal~\cite{rahman2018zero} introduce the max-margin loss into Faster R-CNN to distinguish different classes by using semantic information. Bansal \etal~\cite{bansal2018zero} use the linear projection to map the proposals from R-CNN to a word-vector and then classify them. They also develop an EM-like method to learn the word-vector for background class in an iteration manner. Zhu \etal~\cite{zhu2019zero} develop the ZS-Yolo based on the one-step Yolo detector. Rahman \etal~\cite{rahman2020improved} propose polarity loss to increase the distance between classes and adopt RetinaNet~\cite{retinaNet} as base detector for ZSD. Li \etal~\cite{li2019zero} use textual descriptions as the extra information and develop an attention mechanism to address ZSD problem. Zhao \etal~\cite{zhao2020gtnet} use Generative Adversarial Networks to synthesize semantic representations for unseen objects to help detect unseen objects. Zhu \etal~\cite{zhu2020don} propose DELO that synthesizes visual features for unseen objects from semantic information and incorporate the detection for seen and unseen objects. Zheng \etal~\cite{zheng2020BLC} boost the ZSD performance with a cascade structure to progressively refine the visual-semantic mapping relationship and propose BLRPN to learn the word-vector for background class. In terms of the background representation, SB~\cite{bansal2018zero} and PL~\cite{rahman2020improved} use the fixed word-vector for ``background" word, DSES~\cite{bansal2018zero} and BLC~\cite{zheng2020BLC} try to learn a more reasonable representation. However, the training methods for DSES~\cite{bansal2018zero} and BLC~\cite{zheng2020BLC} are multi-step processes. The background word-vector learned from these non-end-to-end training processes is a local optimum result and it is used as a fixed representation. In our method, we learn the background word-vector in a full end-to-end jointly training manner, which can get an optimal result. In addition, benefiting from our Synchronized Background Strategy, the background word-vector learned from our method is dynamically adaptive compared to the previous fixed form in the other methods, which can greatly improve the recall rate for unseen instances. We note that Bucher \etal~\cite{bucher2019zero} propose the ZSS task that performs semantic segmentation of the entire image, which is different from our method aiming to do instance segmentation for each individual unseen instance.

\begin{figure}[tbp]
\begin{center}
\includegraphics[width=1.0\linewidth]{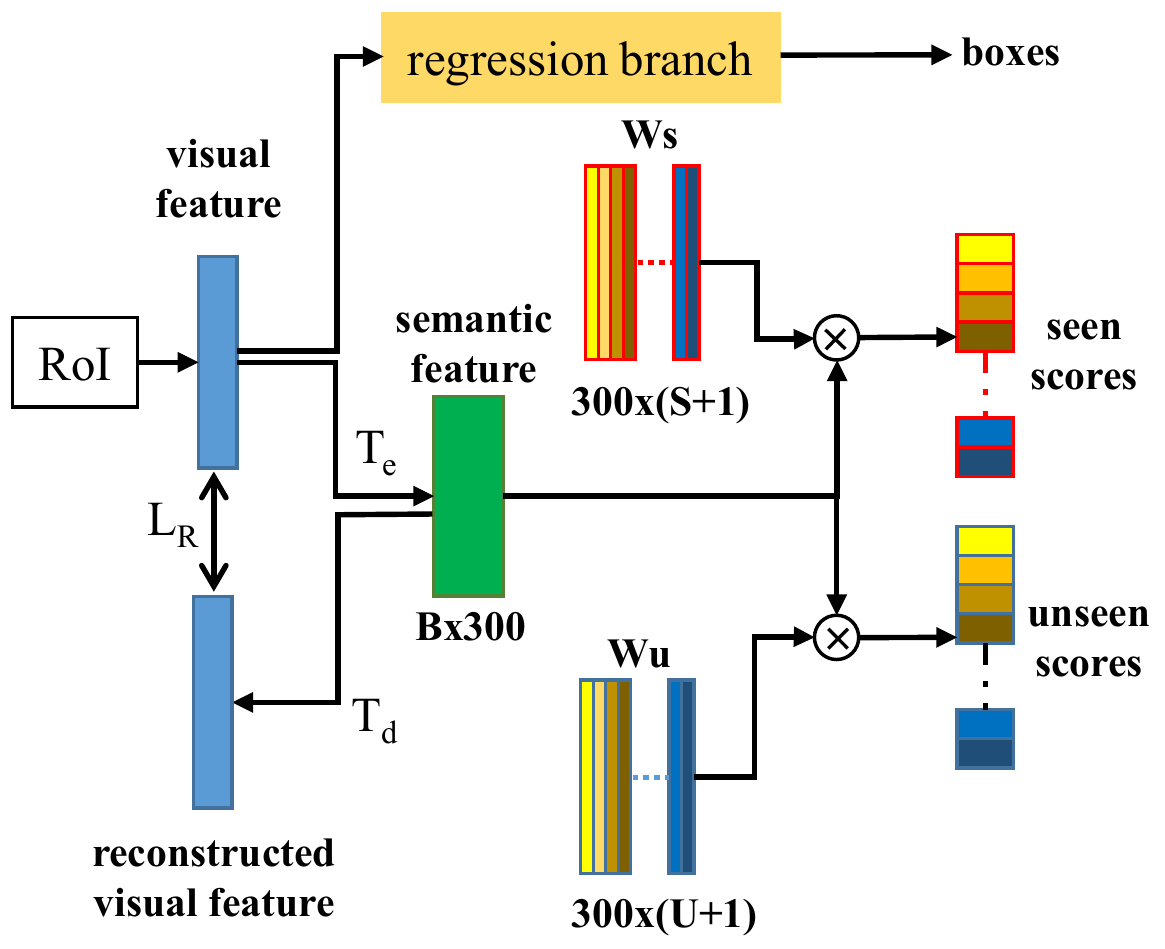}
\caption{The details for zero-shot detector. It is trained in an encoder-decoder manner and we only use the encoder $T_e$ in testing process. $W_s$ is the word-vectors of all seen classes and background class. $W_u$ is the word-vectors of all unseen classes and background class. $S$ is the number of seen classes and $U$ is the number of unseen classes. Each class has a 300-dimensional word-vector. $B$ is batch size.}
\label{fig:zero-shot-detector}
\end{center}
\vspace{-0.5cm}
\end{figure}
\begin{figure*}[tbp]
\begin{center}
\includegraphics[width=0.8\linewidth]{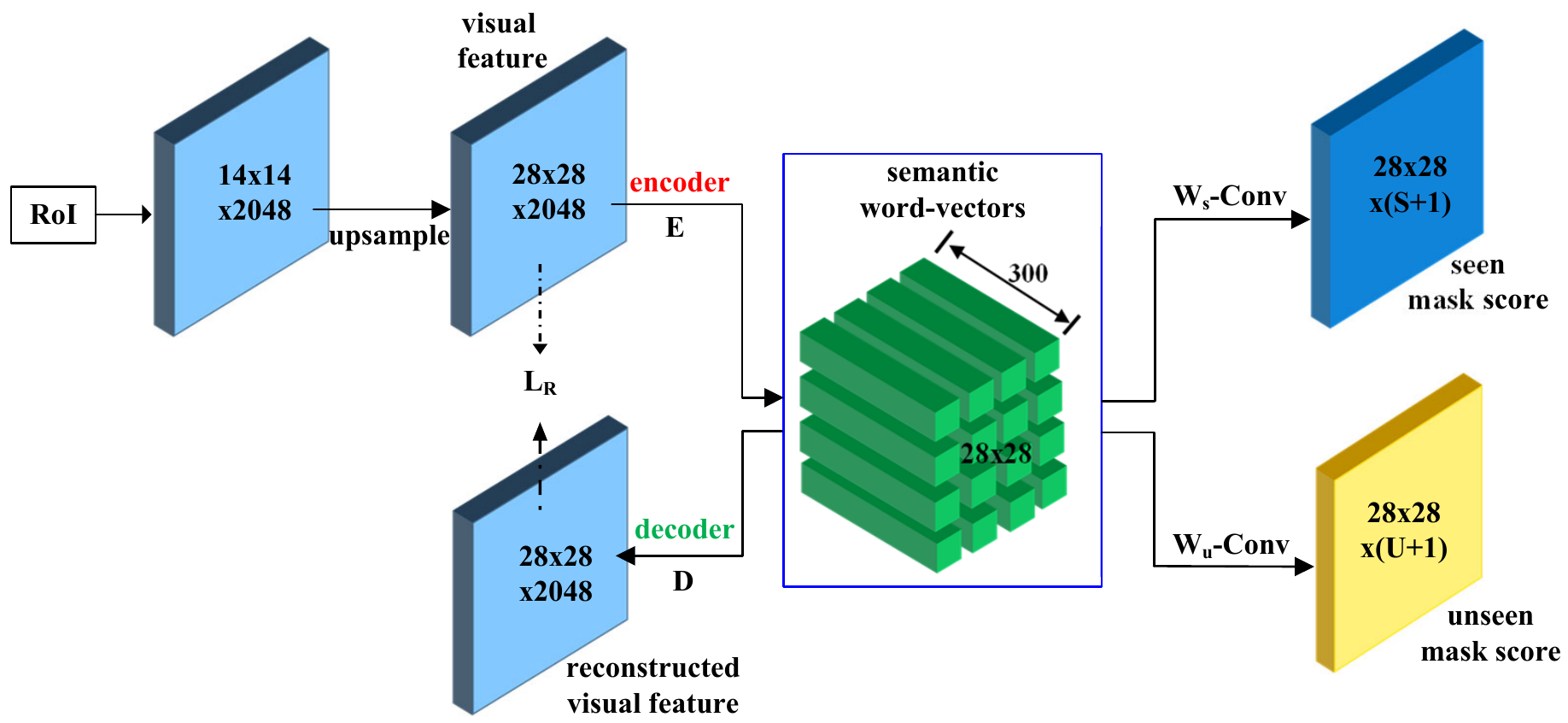}
\caption{Our Semantic Mask Head is an encoder-decoder structure. In training, we use the encoder $E$ to encode the visual feature into the semantic word-vectors. Then we adopt the decoder $D$ to decode the semantic word-vectors back to reconstructed visual feature and use the loss function $\mathcal{L}_R$ to minimize the difference between the two visual features. $D$ is be removed in inference. $W_{s}$-Conv and $W_{u}$-Conv are both fixed convolutional layers and we use them to perform pixel-by-pixel convolution on the semantic word-vectors to get the seen and unseen classes instance segmentation results.}
\label{fig:semantic-mask-figure}
\end{center}
\vspace{-0.5cm}
\end{figure*}
\section{Methodology}
\subsection{Problem Definition}
The setting of zero-shot instance segmentation problem in this paper is described below. Suppose we are given images and word-vectors from two non-overlapping sets of classes: seen classes $C_s$ and unseen classes $C_u$. The training set $D_{train}$ is constructed from $C_s$ which consists of images $x_s$ and seen classes word-vectors $w_s$. The testing set $D_{test}$ is build from $C_s$ and $C_u$, which consists of images $x$ and word-vectors $w$, where the seen and unseen instances may appear at the same image. In training, we use $\theta=\arg\max\limits_{\theta}\sum_{i=1}^{D_{train}}\log p(y_{si}\in C_{s}\mid x_{si}, w_{s}, \theta)$ to train the network, where we use the $D_{train}$ which only contains the instance for seen classes to optimize the parameters $\theta$ of the network. For inference, the target is changed to $\arg\max\limits_{\theta} \sum_{i=1}^{D_{test}} \log p(y_{si}\in C_s, y_{ui}\in C_u \mid x_{i},w, \theta)$, which means using the trained network $\theta$ to get the precisely instance segmentation results for both the seen and unseen classes. In summary, we learn the $\theta$ from seen instance and use it to inference unseen instances.

\subsection{Zero-Shot Instance Segmentation}
We propose an end-to-end network that adopts the semantic word-vector to detect and segment unseen instances. The idea of visual-semantic mapping relationship is embodied in the entire network, including the BA-RPN, Zero-shot Detector and SMH. Fig~\ref{fig:whole-framework} shows the whole architecture of our network. We build our zero-shot detector based on Faster R-CNN with the visual-semantic alignment. Then, we introduce our SMH into zero-shot detector to enable the instance segmentation for unseen classes by learning visual-semantic relationship with an encoder-decoder structure. In addition, we develop BA-RPN and Sync-bg to learn a dynamically adaptive word-vector for background class.

\subsubsection{Zero-Shot Detector}
\label{sec:zero-shot detector}
The main idea for our Zero-Shot Detector is learning the relationship between visual and semantic concepts from seen classes data and transferring it to detect unseen objects. To this end, we replace the classification branch in Faster R-CNN with a new semantic-classification branch. Fig~\ref{fig:zero-shot-detector} indicates its details. The semantic-classification branch is an encoder-decoder structure. In that, we use $T_e$ to encode the visual feature for the input RoI into the semantic feature and use $T_d$ to decode the semantic feature back into visual feature during training. $\mathcal{L}_R$ is the reconstruction loss function to reduce the reconstruction error for visual feature and reconstructed visual feature, which is formulated in Eq~\ref{con:eq3}. The decoder module and reconstruction loss can push the network to learn a more discriminative visual-semantic alignment. In inference, $T_d$ is be removed and we can get the scores for seen and unseen classes by performing the matrix multiplication of the matrices semantic feature and $W_s$, semantic feature and $W_u$, respectively.
\subsubsection{Semantic Mask Head}
\label{sec:smh}
In order to do instance segmentation for unseen instance, we focus on how to segment them using the visual-semantic mapping relationship. We propose Semantic Mask Head to learn this relationship and transform it from seen classes to segmenting the unseen instances. The details for our Semantic Mask Head are illustrated in Fig~\ref{fig:semantic-mask-figure}. The overall architecture is an encoder-decoder structure. The encoder module $E$ is a single $1\times 1$ convolutional layer structure, which encodes the visual features into semantic space, and then we can get the segmentation result from these semantic word-vectors. Considering that this single forward encoder structure is not enough to learn a tight visual-semantic alignment, we develop a decoder structure $D$ with a reconstruction loss function $\mathcal{L_R}$ to further improve the quality of the mapping relationship between visual features and semantic word-vectors. Our decoder module decodes the semantic word-vectors back to visual features and we use $\mathcal{L_R}$ to minimize the difference between the reconstructed visual feature and the original visual feature. For this purpose, the decoder module has a symmetrical structure relative to the encoder module, which also includes a single $1\times 1$ convolutional layer, in which the number of input and output channels are opposite to $E$. We use a 300-dimensional word-vector as the semantic representation for each class, so $E$ is responsible for converting the input visual feature into the semantic feature with a channel dimension of 300. In this $300\times28\times28$ semantic feature tensor, each channel represents a dimension of the word-vector and each $300\times1$ element is a word-vector. To get the classification score for each element, we need to calculate the similarity between the word-vector of each element and the word-vectors of all seen and unseen classes, so as to find the closest category. To this propose, we add a classification module after the encoder. This module includes two branches, one for seen classes and the other one for unseen classes. In it, $W_{s}$-Conv indicates a fixed $1\times1$ convolutional layer and we adopt the word-vectors of all seen classes and background class $W_s$ as the weight of it. $W_{u}$-Conv is also a fixed $1\times1$ convolutional layer with the weight of $W_u$, which indicates the word-vectors of all unseen classes and background class. The $W_s$ and $W_u$ are the same in our zero-shot detector. We merge the results of seen and unseen classes through scores.

We use mse loss for the reconstruction loss $\mathcal{L}_{R}$ in zero-shot detector and SMH. As shown in Equation~\ref{con:eq3}, we measure the mean square error (squared $\ell_2$ norm) between each element in the original visual feature $O$ and the reconstructed visual feature $R$. 
\begin{equation}
\label{con:eq3}
\begin{split}
    L_{R} = \sum_{i=1}^{F}(O_i-R_i)^2
\end{split}
\end{equation}
\subsubsection{BA-RPN and Synchronized Background}
\label{sec:barpn and sync-bg}
\begin{figure}[tbp]
\begin{center}
\includegraphics[width=1.0\linewidth]{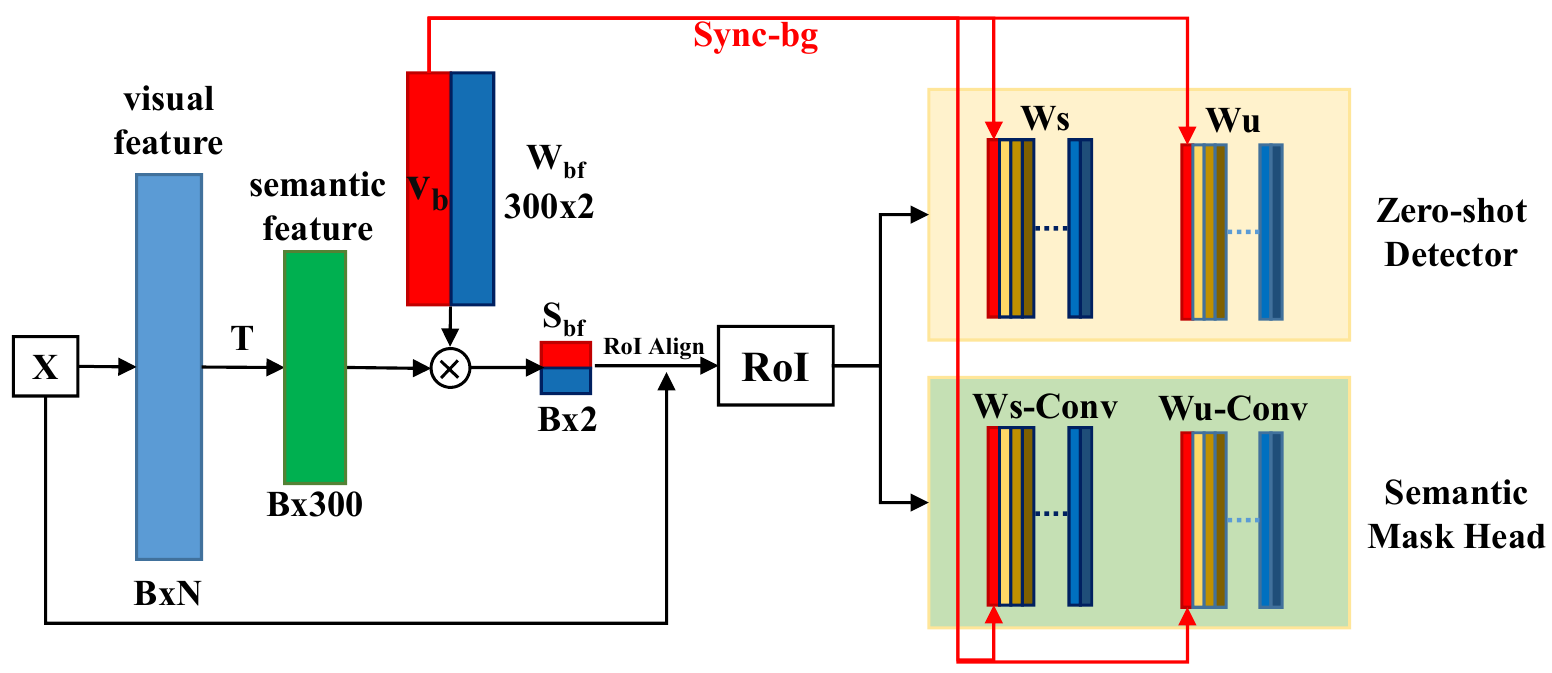}
\caption{This is the details for BA-RPN and Sync-bg. In BA-RPN, we use $T$ to transform the $B\times N$ dimensional visual feature to the $B\times300$ dimensional semantic feature, where $B$ is batch size and $N$ is the dimension of the visual feature. Then we use $W_{bf}$ which contain the background class word-vector $v_b$ to get the foreground-background classification score $S_{bf}$. We use the Sync-bg to synchronize the $v_b$ in the $W_s$ and $W_u$ for Zero-shot Detector, the $W_s$-Conv and $W_u$-Conv for Semantic Mask Head.}
\label{fig:BARPN-sync-bg}
\end{center}
\vspace{-0.5cm}
\end{figure}

In our zero-shot detector, word-vectors $W_s$ and $W_u$ for seen and unseen classes are used as the weights value for the fixed FC layers to classify the objects into $n+1$ categories. Where $n$ is the number of seen or unseen classes and the $1$ represents the background class. In our Semantic Mask Head, $W_s$ and $W_u$ are used again as the weights value of $W_s$-Conv and $W_u$-Conv to classify the pixels into $n+1$ categories, which means that our Semantic Mask Head also classifies each pixel to background class or other target classes. From the above discussions, we can learn that the word-vector of background class in $W_s$ and $W_u$ directly affect the classification result of the background class. However, the existing word-vector for background class is learned from large-scale text data without using the visual information, and can not effectively represents complex background class. To this end, we propose the Background Aware Region Proposal Network (BA-RPN), which introduces the visual-semantic learning process into the original RPN to learn a more reasonable word-vector for background class from images. The architecture of BA-RPN is indicated in Fig~\ref{fig:BARPN-sync-bg}, which uses an FC layer $T$ to transform the input visual features into semantic features. We use a $300\times2$ dimensional FC layer $W_{bf}$ to get the background-foreground binary classification score from the input visual feature. The weight for $W_{bf}$ is a $300\times2$ vectors, which represents the two word-vectors for foreground and background to learn. $W_{bf}$ will be optimized during training so that we can learn a new word-vector $v_b$ for background class.

Now we have a new word-vector for background class which can be used to replace the original one in our zero-shot detector and SMH. However, we believe this is still not the best way to solve the problem of background representation. The background class has different forms in different images, while the background word-vector learned from BA-RPN is still a fixed one. The training process for BA-RPN and the whole ZSI framework are divided, which leads to the background learning limited to BA-RPN, and the benefits of zero-shot detector and SMH are not well utilized. To fix these issues, we propose the Synchronized Background Strategy and use this synchronous operation in training and inference. Algorithm~\ref{alg:sync-bg} indicates the details for our Sync-bg in training process. In each training step, we first forward the visual features from backbone into BA-RPN and get the word-vector $v_b$ for background class and the visual features for each proposal. Then we update the $W_s$ in zero-shot detector and the $W_s$-Conv in SMH by replacing the word-vector for background with $v_b$. After we forward the visual features into zero-shot detector and SMH, we calculate the loss and propagate the gradient back to update all trainable parameters, including the $v_b$ in BA-RPN. These end-to-end jointly training process can learn a more reasonable $v_b$. In inference, for an input image, BA-RPN can adaptively output the corresponding background word-vector to it and the Synchronized Background Strategy synchronizes this adaptive background word-vector to update the $W_s$, $W_u$ in zero-shot detector and $W_s$-Conv, $W_u$-Conv in SMH. Using this dynamic adaptive background word-vector in our zero-shot detector and SMH significantly improves the performance.
 \begin{algorithm}[tbp]  
  \caption{Synchronized Background in Training.}
  \label{alg:sync-bg}  
  \begin{algorithmic}[1]  
    \Require The visual features from backbone network $x$; 
    \ForAll {training iterative steps}
    \State Forward $x$ into BA-RPN to get background word-vector $v_b$ and features $x_p$ for all proposals;
    \label{code:fram:extract bg word-vector}
    \State Update the $W_s$ in zero-shot detector with $v_b$;
    \label{code:fram:update bg in det}
    \State Update the $W_s$-Conv in SMH with $v_b$;
    \label{code:fram:update bg in mask}
    \State Forward $x_p$ through zero-shot detector and SMH;
    \label{code:fram:forward}  
    \State Calculate loss and back propagate gradient to update $v_b$ in BA-RPN.;
    \label{code:fram:loss back}  
    \EndFor
  \end{algorithmic}
\end{algorithm} 

\begin{table*}[tbp]
\begin{center}
\caption{Effects of each component in our work. Results are reported on 48/17 split and 65/15 split of MS-COCO, respectively.}
\label{table:component-wise}
\resizebox{0.8\textwidth}{!}{
\begin{tabular}{c|ccccccccccc}
\toprule
\multirow{3}{*}{ } & \multirow{3}{*}{ZSD} & \multicolumn{2}{c}{\multirow{3}{*}{SMH}} & \multirow{3}{*}{Det Decoder} & \multirow{3}{*}{BA-RPN \& Sync-bg} & \multicolumn{4}{c}{ZSI} & ZSD \\
\cmidrule(lr){7-10} \cmidrule(lr){11-11}
~ & ~ & ~ & ~ & ~ & ~ & \multicolumn{3}{c}{Recall@100} & mAP  & Recall@100 \\
\cmidrule(lr){3-4} \cmidrule(lr){7-9}  \cmidrule(lr){10-10} \cmidrule(lr){11-11}
~ & ~ & Encoder & Decoder & ~ & ~ & 0.4 & 0.5 & 0.6 & 0.5 & 0.5 \\
\cmidrule(lr){1-11}
\multirow{4}{*}{\rotatebox{90}{48/17}} & \checkmark &  &  & & & - & - & - & - & 47.2 \\
~ & \checkmark &  \checkmark &  & & & 43.8 & 38.5 & 32.7 & 7.5 & 48.1 \\
~ & \checkmark &  \checkmark  &  \checkmark &  & & 46.1 & 41.2 & 35.5 & 8.4 & 48.6 \\
~ & \checkmark &  \checkmark &  \checkmark & \checkmark & & 46.8 & 41.8 & 35.9 & 8.6 & 49.3 \\
~ & \checkmark &  \checkmark &  \checkmark & \checkmark & \checkmark & \textbf{50.3} & \textbf{44.9} & \textbf{38.7} & \textbf{9.0} &  \textbf{53.9} \\
\bottomrule
\noalign{\smallskip}
\multirow{4}{*}{\rotatebox{90}{65/15}} & \checkmark &  &  & & & - & - & - & - & 52.9 \\
~ & \checkmark &  \checkmark &  & & & 48.9 & 42.6 & 35.5 & 9.1 & 53.4 \\
~ & \checkmark &  \checkmark  &  \checkmark &  & & 51.7 & 45.8 & 38.7 & 10.1 & 54.1 \\
~ & \checkmark &  \checkmark &  \checkmark & \checkmark & & 52.4 & 47.0 & 40.5 & 10.3 & 55.0 \\
~ & \checkmark &  \checkmark &  \checkmark & \checkmark & \checkmark & \textbf{55.8} & \textbf{50.0} & \textbf{42.9} & \textbf{10.5} &  \textbf{58.9} \\
\bottomrule
\end{tabular}
}
\end{center}
\vspace{-0.5cm}
\end{table*}
\section{Loss Function.}
As shown in Eq~\ref{con:eq4}, the whole loss function $\mathcal{L_{ZSI}}$ for our network has three components: the loss of BA-RPN $\mathcal{L_{BA}}$, the loss of zero-shot detector $\mathcal{L_{ZSD}}$ and the loss for SMH $\mathcal{L_{SMH}}$. $\mathcal{L_{BA}}$ includes foreground-background classification cross-entropy loss and the smooth $\ell{1}$ regression loss. $\mathcal{L_{ZSD}}$ is consist of the cross-entropy classification loss, the smooth $\ell{1}$ regression loss and the reconstruction loss. $\mathcal{L_{SMH}}$ includes a per-pixel binary classification loss and the reconstruction loss. $CE$ is the cross-entropy loss and $BCE$ is binary cross-entropy loss. $ \lambda_{ZSD}$ and $\lambda_{SMH}$ are hyperparameters.
\begin{equation}
\label{con:eq4}
\begin{split}
& \mathcal{L_{ZSI}} = \mathcal{L_{BA}} + \mathcal{L_{ZSD}} + \mathcal{L_{SMH}} \\
& \mathcal{L_{BA}} = \ell_{1}(\mathbf{r}, \widehat{\mathbf{r}}) + CE(c, \widehat{c}) \\
& \mathcal{L_{ZSD}} =\ell_{1}(\mathbf{r}, \widehat{\mathbf{r}}) + CE(c, \widehat{c}) + \lambda_{ZSD}\mathcal{L}_{R}(O,R)\\
& L_{SMH} = BCE(c, \widehat{c}) + \lambda_{SMH}\mathcal{L}_{R}(O,R)\\
\end{split}
\end{equation}

\section{Experiments}
\subsection{Datasets}
Considering that instance segmentation task is always regarded as a downstream task for object detection, we develop the datasets for ZSI by following the previous ZSD works~\cite{bansal2018zero,rahman2020improved}. We use MS-COCO as the basic dataset because it is widely used as the common benchmark for object detection and instance segmentation. We choose the 2014 version for MS-COCO, because this version has more data in validation set than the 2017 version, so we can have more data to evaluate our method. In order to construct the benchmark, we give two division methods of seen and unseen classes: 48/17 split and 65/15 split, which means we split the MS-COCO into 48 seen classes with 17 unseen classes and 65 seen classes with 15 unseen classes. For the training set, we first select all images which contains seen classes objects from the training set of MS-COCO. Then we remove the image in this selected set if it contains any unseen object to guarantee that the unseen objects will not be observed by the network during training. For test set, we extract all images which contains unseen objects from the validation set for MS-COCO. The images in our test set may contain seen and unseen objects together. A detailed description of the datasets can found in the appendix.

\subsection{Evaluation Protocol}
We evaluate the performance for ZSI with two settings: the ZSI setting and the generalize zero-shot instance segmentation (GZSI) setting. When using ZSI setting, the network only needs to predict the results for unseen instances. For GZSI setting, the results for seen and unseen classes need to be predicted together. GZSI is closer to the situation in real world because the seen and unseen instances may appear at the same time. Referring to the previous works of ZSD~\cite{bansal2018zero, rahman2020improved, zheng2020BLC, zhao2020gtnet}, we use Recall@100 across different IoU thresholds (0.4, 0.5, 0.6) as the main metric and the 100 means we select the top 100 score results for evaluation. In addition, we also give the results of mean average precision (IoU threshold is 0.5) as a reference.

\subsection{Implement Details}
We adopt word2vec~\cite{mikolov2013distributed} as our semantic word-vector and normalize it with a $\ell_2$ normalization. In zero-shot detector, $T_e$ is an FC layer with output dimension 300 and $T_d$ is a 2048 output dimensional FC layer. $W_s$ and $W_u$ are two fixed FC layer which are not updated during training. In SMH, $E$ and $D$ are two $1\times1$ convolutional layers. $W_s$-Conv and $W_u$-Conv are both fixed $1\times1$ convolutional layers and the weight for them are the word-vectors for seen and unseen classes. The weight for reconstruction loss function in zero-shot detector and SMH are both set to 0.5.
\begin{table}[ht]
\caption{Comparison of our method with the previous state-of-the-art ZSD works on two splits of COCO. Seen/Unseen refers to the split of datasets. Our method significantly surpasses all other works.}
\vspace{-0.2cm}
\label{table:zsd benchmark result}
\begin{center}
\resizebox{0.9\linewidth}{!}{
\begin{tabular}{cccccc}
\toprule
\multirow{2}{*}{Method} & \multirow{2}{*}{Seen/Unseen} & \multicolumn{3}{c}{Recall@100} & mAP \\
\cmidrule(lr){3-5} \cmidrule(lr){6-6}
~ & ~ & 0.4 & 0.5 & 0.6 & 0.5 \\
\midrule
SB~\cite{bansal2018zero} & 48/17 & 34.46 & 22.14 & 11.31 & 0.32 \\
DSES~\cite{bansal2018zero} & 48/17 & 40.23 & 27.19 & 13.63 & 0.54 \\
TD~\cite{li2019zero} & 48/17 & 45.50 & 34.30 & 18.10 & - \\
PL~\cite{rahman2020improved} & 48/17 & - & 43.59 & - & 10.1 \\ 
Gtnet~\cite{zhao2020gtnet} & 48/17 & 47.30 & 44.60 & 35.50 & - \\ 
DELO~\cite{zhu2020don} & 48/17 & - & 33.50 & - & 7.6 \\ 
BLC~\cite{zheng2020BLC} & 48/17 & 49.63 & 46.39 & 41.86 & 9.9
\\
\hline
\textbf{ZSI} & 48/17 &  \textbf{57.4} &  \textbf{53.9} &  \textbf{48.3} & \textbf{11.4} \\
\bottomrule
PL~\cite{rahman2020improved} & 65/15 & - & 37.72 & - & 12.4  \\
BLC~\cite{zheng2020BLC} & 65/15 & 54.18 & 51.65 & 47.86 & 13.1 \\
\hline
\textbf{ZSI} & 65/15 & \textbf{61.9} & \textbf{58.9} & \textbf{54.4} & \textbf{13.6} \\
\bottomrule
\end{tabular}}
\end{center}
\vspace{-0.5cm}
\end{table}
\subsection{Component-wise Analysis}
We investigate the contributions of the main components for our method. ``ZSD" means our Zero-shot Detector, ``SMH" denotes the Semantic Mask Head, ``Encoder" and ``Decoder" under the ``SMH" mean the encoder and decoder module in Semantic Mask Head, respectively. ``Det Decoder" represents adding the decoder module into zero-shot detector and ``BA-RPN\&Sync-bg" indicates the BA-RPN and Synchronized Background Strategy. The results for 48/17 and 65/15 splits are shown in Table~\ref{table:component-wise}, respectively. Compared with the baseline, our method brings 6.4\% and 7.4\% improvement for ZSI, 6.7\% and 7.2\% for ZSD in terms of Recall@100 for 48/17 and 65/15 splits, respectively.
\begin{table}[tbp]
\caption{This table shows the performances in Recall@100 and mAP (IoU threshold=0.5) for our method and other state of the art over GZSD task. HM denotes the harmonic average for seen and unseen classes.}
\vspace{-0.5cm}
\label{table:gzsd result}
\begin{center}
\resizebox{\linewidth}{!}{
\begin{tabular}{cccccccc}
\toprule
\multirow{2}{*}{Method} & \multirow{2}{*}{Seen/Unseen} & \multicolumn{2}{c}{seen} & \multicolumn{2}{c}{unseen} & \multicolumn{2}{c}{HM} \\
\cmidrule(lr){3-4} \cmidrule(lr){5-6} \cmidrule(lr){7-8}
~ & ~ & mAP & Recall & mAP & Recall & mAP & Recall \\
\midrule
DSES~\cite{bansal2018zero} & 48/17 & - & 15.02 & - & 15.32 & - & 15.17 \\
PL~\cite{rahman2020improved} & 48/17 & 35.92 & 38.24 & 4.12 & 26.32 & 7.39 & 31.18 \\
BLC~\cite{zheng2020BLC} & 48/17 & 42.10 & 57.56 & 4.50 & 46.39 & 8.20 & 51.37 \\
\hline
\textbf{ZSI} & 48/17 & \textbf{46.51} & \textbf{70.76} &  \textbf{4.83} & \textbf{53.85} & \textbf{8.75} & \textbf{61.16} \\
\bottomrule
PL~\cite{rahman2020improved} & 65/15 & 34.07 & 36.38 & 12.40 & 37.16 & 18.18 & 36.76 \\
BLC~\cite{zheng2020BLC} & 65/15 & 36.00 & 56.39 & 13.10 & 51.65 & 19.20 & 53.92  \\
\hline
\textbf{ZSI} & 65/15 & \textbf{38.68} & \textbf{67.11} & \textbf{13.60} & \textbf{58.93} & \textbf{20.13} & \textbf{62.76}  \\
\bottomrule
\end{tabular}}
\end{center}
\vspace{-0.8cm}
\end{table}
\subsection{Comparison with Other ZSD Methods}
We compare our method with the state-of-the-art zero-shot detection approaches on two split benchmarks in Table~\ref{table:zsd benchmark result} over ZSD setting. We can observe that: (i) for 48/17 split, we compare our approaches with SB~\cite{bansal2018zero}, DSES~\cite{bansal2018zero}, TD~\cite{li2019zero}, PL~\cite{rahman2020improved}, Gtnet~\cite{zhao2020gtnet}, DELO~\cite{zhu2020don} and BLC~\cite{zheng2020BLC}. Our method surpasses all of them, brings up to 36.99\% and 11.08\% gain in terms of Recall@100 and mAP; (ii) for 65/15 split, compared with PL~\cite{rahman2020improved} and BLC~\cite{zheng2020BLC}, our method still gets the best performance and brings 21.18\% gain for Recall@100 and 1.2\% improvement for mAP. In addition, we also compare the performance under GZSD setting~\cite{bansal2018zero} in Table~\ref{table:gzsd result}, which require our network to predict seen and unseen at the same time. We can learn that our method obtains up to 46\% and 26\% improvement for seen and unseen classes in Recall@100 compared with DSES~\cite{bansal2018zero}, PL~\cite{rahman2020improved} and BLC~\cite{zheng2020BLC}.
\begin{table}[tbp]
\caption{This table shows the performances in Recall@100 and mAP (IoU threshold=0.5) for our method over GZSI task. HM denotes the harmonic average for seen and unseen classes.}
\vspace{-0.5cm}
\label{table:gzsi result}
\begin{center}
\resizebox{\linewidth}{!}{
\begin{tabular}{cccccccc}
\toprule
\multirow{2}{*}{Method} & \multirow{2}{*}{Seen/Unseen} & \multicolumn{2}{c}{seen} & \multicolumn{2}{c}{unseen} & \multicolumn{2}{c}{HM} \\
\cmidrule(lr){3-4} \cmidrule(lr){5-6} \cmidrule(lr){7-8}
~ & ~ & mAP & Recall & mAP & Recall & mAP & Recall \\
\midrule
baseline & 48/17 & 41.11 & 62.30 & 3.01 & 38.14 & 5.61 & 47.31 \\
\hline
\textbf{ZSI} & 48/17 & \textbf{43.04} & \textbf{64.48} &  \textbf{3.65} & \textbf{44.90} & \textbf{6.73} & \textbf{52.94} \\
\bottomrule
baseline & 65/15 & 34.43 & 60.59 & 9.06 & 43.91 & 14.35 & 50.92 \\
\hline
\textbf{ZSI} & 65/15 & \textbf{35.75} & \textbf{62.58} & \textbf{10.47} & \textbf{49.95} & \textbf{16.20} & \textbf{55.56}  \\
\bottomrule
\end{tabular}}
\end{center}
\vspace{-0.5cm}
\end{table}
\begin{table}[tbp]
\begin{center}
\caption{Effectiveness for BA-RPN and Synchronized Background Strategy. The Recall@100 results for ZSI and ZSD are reported on 48/17 split and 65/15 split of MS-COCO, respectively.}
\label{table:BA-RPN sync-bg}
\resizebox{\linewidth}{!}{
\begin{tabular}{c|ccccc}
\toprule
\multirow{2}{*}{ } & \multirow{2}{*}{Method} & \multicolumn{3}{c}{ZSI} & ZSD \\
\cmidrule(lr){3-5} \cmidrule(lr){6-6}
~ & ~ & 0.4 & 0.5 & 0.6 & 0.5\\
\cmidrule(lr){1-6}
\multirow{4}{*}{\rotatebox{90}{48/17}} & RPN & 43.8 & 38.5 & 32.7 & 48.1\\
~ & BA-RPN & 43.1 & 37.9 & 31.4 & 48.9\\
~ & BA-RPN\&Sync-bg in mask & 44.1 & 38.8 & 32.7 & 48.6\\
~ & BA-RPN\&Sync-bg in det & 46.7& 40.9 & 35.0 & 51.9 \\
~ & \textbf{BA-RPN\&Sync-bg} & \textbf{47.9} & \textbf{42.1} & \textbf{35.4} & \textbf{54.2} \\
\bottomrule
\noalign{\smallskip}
\multirow{4}{*}{\rotatebox{90}{65/15}} & RPN & 48.9 & 42.6 & 35.5 & 53.4 \\
~ & BA-RPN & 49.0 & 42.4 & 35.4 & 55.3 \\
~ & BA-RPN\&Sync-bg in mask & 49.0 & 42.5 & 35.4 & 55.0\\
~ & BA-RPN\&Sync-bg in det & 51.9 & 45.3 & 37.9 & 57.6\\
~ & \textbf{BA-RPN\&Sync-bg} & \textbf{53.7} & \textbf{47.4} & \textbf{39.9} & \textbf{58.5}\\
\bottomrule
\end{tabular}
}
\end{center}
\vspace{-0.5cm}
\end{table}
\subsection{Generalized Zero-Shot Instance Segmentation}
For the generalized zero-shot instance segmentation setting, we need to segment the instances for seen and unseen classes as the same time, which means GZSI task is more challenging than ZSI. We build a baseline that has no decoder in detector and SMH, no BARPN, and no Sync-bg. We report the GZSI results in Table~\ref{table:gzsi result}. We can learn that our method brings up to 5.61\% for mAP and 11.72\% for Recall@100  improvements on the harmonic average for seen and unseen classes over GZSI setting compared with the baseline.
\begin{figure*}[tbp]
\centering
\includegraphics[width=0.8\linewidth]{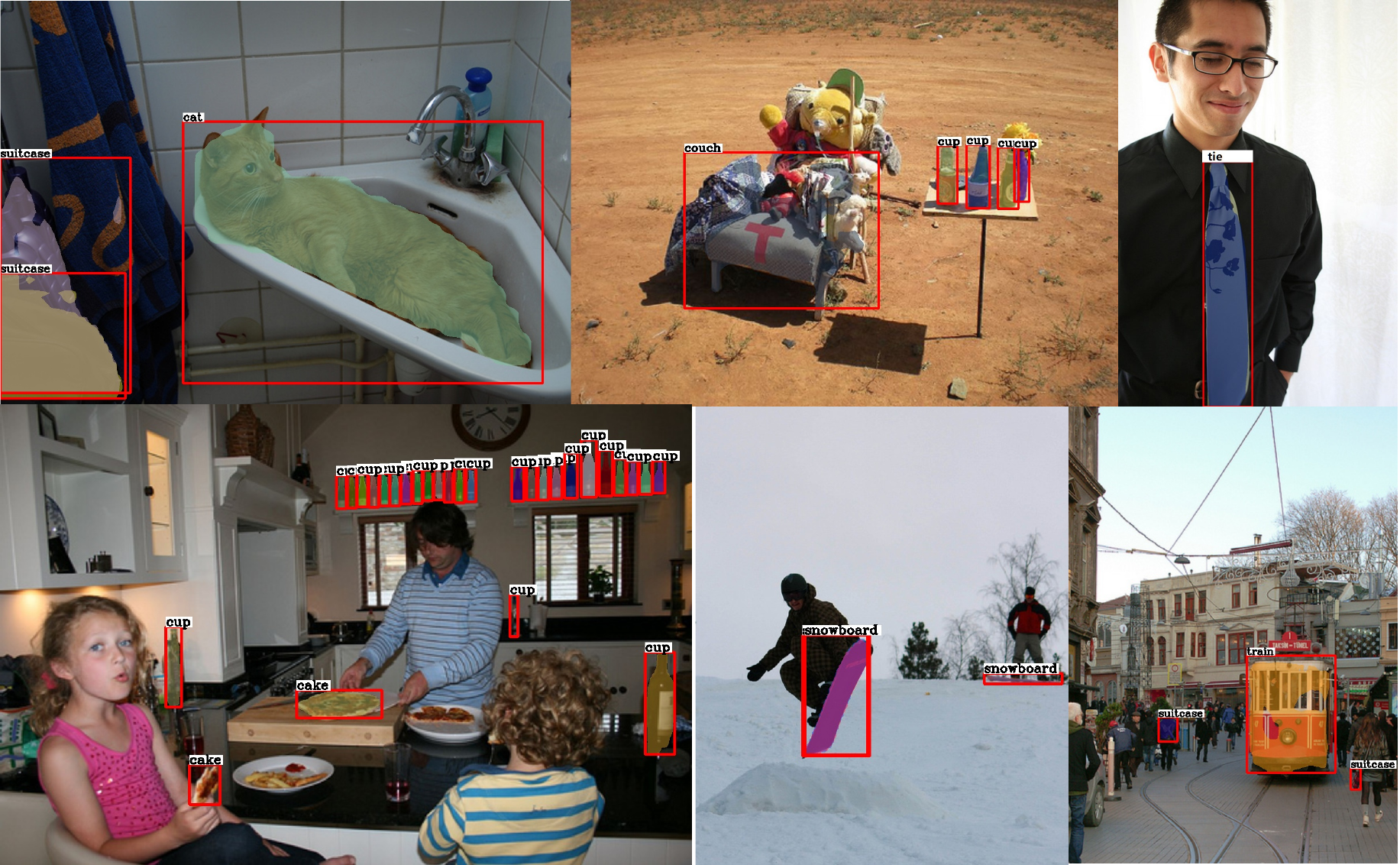}
\caption{Examples for the results of zero-shot instance segmentation from our method. All of these instances belong to unseen classes.}
\label{fig:demo}
\vspace{-0.2cm}
\end{figure*}
\subsection{Ablation Studies for BA-RPN and Sync-bg}
We report the experimental results about the BA-RPN and Sync-bg to discuss the effectiveness for them. We build a baseline network by only combining the zero-shot detector and the encoder module in SMH. We conducted a series of ablation experiments on this basis, see Table~\ref{table:BA-RPN sync-bg}. ``RPN" means above baseline, ``BA-RPN" is replacing RPN with our Background Aware RPN, ``BA-RPN\&Sync-bg in det" denotes we only use the Sync-bg strategy in BA-RPN and zero-shot detector, ``BA-RPN\&Sync-bg in mask" indicates that we use the Sync-bg strategy in BA-RPN and SMH, ``BA-RPN\&Sync-bg" represents that we use Sync-bg strategy through BA-RPN, zero-shot detector and SMH. From these results, we can learn that: (i) using BA-RPN alone is even worse than the original RPN without Sync-bg strategy, it degrades ZSI performance by 0.4\% to 0.6\%, which verifies that using BA-RPN alone does not bring benefits; (ii) we need synchronize the background word-vector learned from BA-RPN to the other components. If we only use Sync-bg in SMH, it does not bring much improvement for ZSI and reduces the ZSD performance. But when we use Sync-bg in detector, we can get a significantly improved for both ZSD and ZSI. We believe that the reason for this phenomenon is that in the prediction process, we first use the zero-shot detector to obtain the bounding boxes, and then perform instance segmentation on these boxes. If we only synchronize the background word-vector in semantic mask head without zero-shot detector, the inconsistency of the background word-vector in zero-shot detector and semantic mask head will decrease ZSI performance. When we maintain the consistency of the background word-vector that synchronize the background word-vector in whole framework, we can get the best performance.

\subsection{The effect of the semantic information}
We add a experiment to explore the effect of the semantic information. The results is shown in Table~\ref{table:trained-wordvector}. We can learn that if one-hot vectors were used for the word vectors, the performance on unseen classes should be equivalent to a random baseline because the knowledge in semantic information is necessary.
\begin{table}[tbp]
\caption{Importance of the semantic information. The semantic information needs to contain prior knowledge.}
\label{table:trained-wordvector}
\resizebox{\linewidth}{!}{
\begin{tabular}{cccc}
\toprule
word-vectors & has knowledge & Recall@100 & mAP\\
\midrule
random baseline & No & 0.2 & 0 \\
\hline
one-hot & No & 0.3 & 0 \\
\hline
word-vec (in our method) & \multirow{2}{*}{Yes} & \multirow{2}{*}{58.9} & \multirow{2}{*}{13.6}\\
trained from unannotated text data & ~ & ~ & ~\\
\bottomrule
\end{tabular}}
\vspace{-0.5cm}
\end{table}
\subsection{Qualitative Results}
To intuitively evaluating the qualitative results, we give some instance segmentation results of unseen instances in Fig~\ref{fig:demo} for our method. We find that our method can precisely detect and segment unseen classes in different situations. For example, our method detects and segments instances under densely packed scenes, \eg, ``cup". It is noteworthy that multiple instances are also segmented from a messy background like ``couch", ``cup" and ``snowboard".
\section{Conclusion}
We introduce a novel zero-shot instance segmentation task and provide the evaluation protocol and benchmarks for it. To tackle this new task, we propose a zero-shot instance segmentation network with Zero-shot Detector, Semantic Mask Head, BA-RPN and Synchronized Background Strategy. By making use of the semantic information and learning more reasonable adaptive background representation, our method outperforms other state-of-the-art zero-shot detection approaches and provide a solid baseline for zero-shot instance segmentation task. Worthwhile and related future work can be spawned from our new task with the above technical contributions.
\section*{Acknowledgement}
The paper is supported by the National Natural Science Foundation of China (NSFC) under Grant No. 61672498.
{\small
\bibliographystyle{unsrt}
\bibliography{egbib}
}

\end{document}